\documentclass{article}
\usepackage{graphicx} % Required for inserting images

\usepackage{authblk} % for affiliation commands
\usepackage{amsmath} % for equations
\usepackage{subfiles} % for subfiles
\usepackage{multirow} % for multirow in table
\usepackage[export]{adjustbox}
\usepackage{colortbl}
\usepackage{xcolor}
\usepackage{setspace} 
\title{Comparing Generalization in Learning with Limited Numbers of Exemplars: Transformer vs. RNN in Attractor Dynamics}
\author{Rui Fukushima\thanks{rui.fukushima@oist.jp}}
\author{Jun Tani\thanks{jun.tani@oist.jp}}
\affil{Okinawa Institute of Science and Technology}
\date{}

\begin{document}

\maketitle

\section*{ABSTRACT}
ChatGPT, a widely-recognized large language model (LLM), has recently gained substantial attention for its performance scaling, attributed to the billions of web-sourced natural language sentences used for training. Its underlying architecture, Transformer, has found applications across diverse fields, including video, audio signals, and robotic movement. %The crucial question this raises concerns the Transformer's generalization-in-learning (GIL) capacity. 
However, this raises a crucial question about Transformer's generalization in learning (GIL) capacity. Is ChatGPT's success chiefly due to the vast dataset used for training, or is there more to the story? To investigate this, we compared Transformer's GIL capabilities with those of a traditional Recurrent Neural Network (RNN) in tasks involving attractor dynamics learning. For performance evaluation, the Dynamic Time Warping (DTW) method has been employed. Our simulation results suggest that under conditions of limited data availability, Transformer's GIL abilities are markedly inferior to those of RNN.
\section*{INTRODUCTION}
Among human competencies, the ability to extract generalized knowledge or skills from a relatively limited set of experiences is particularly fascinating \cite{shadmehr1994adaptive,meltzoff1995understanding,mckeough2013teaching}. 
Enhancing this ability, which is known as generalization in learning (GIL), in machines has been a major objective in machine learning and AI \cite{neyshabur2017exploring,zhang2021understanding}. 
The crux of GIL lies in extracting the essential structure underlying a dataset, while disregarding contaminating noise, rather than memorizing each data point individually.
 
Deep learning, with its impressive scalability, has recently attracted great attention \cite{Devlin2018, Brown2020}. The capability of Transformer \cite{Vaswani2017}, ChatGPT's core engine, to correctly answer questions across a diverse array of knowledge domains far surpasses that of the average human. However, a key concern is that Transformer's remarkable performance is reliant on billions of training sentences—a significant divergence from human learning, which requires far fewer experiences to develop knowledge or skills \cite{Brown2020, Frank2023}. % Despite Transformer's many successes, it fails to generalize in many simple tasks that recurrent models handle easily. This is attributed to its lack of ability to acquire inductive bias \cite{Tran2018, Dehghani2019, Abnar2020}. 
Against this background, we propose a hypothesis: while the Transformer excels in scaling when fed vast amounts of data, its generalization capability is considerably weaker than those of other neural network models, especially the Recurrent Neural Network (RNN), when working with a small number of exemplars. 

Previous research has undertaken comparative evaluations of performance between Transformer and RNN architectures across various tasks. For instance, Tran et al. \cite{Tran2018} focused on tasks involving hierarchical structure modeling, such as subject-verb agreement \cite{Linzen2016}, and emphasized the importance of recurrency. Dehghani et al. \cite{Dehghani2018} employed the bAbi question-answering dataset \cite{Weston2015} and suggested the efficacy of incorporating an adequate recurrent inductive bias for several algorithmic and language understanding tasks. While these studies employed relatively simple tasks, the number of training samples employed in these studies still surpasses the number required for human learning, which requires a few experiences to develop knowledge and skills. Therefore, in contrast to these prior works, we provide a simple attractor dynamics learning task to compare GIL in extracting essential structure from a limited set of exemplars. Importantly, our approach focuses on the generative model and involves training the Transformer model without extensive datasets, allowing us to assess their inherent generalization capabilities.

\section*{RESULTS}

\begin{figure}[t]
    \includegraphics[scale = 0.35]{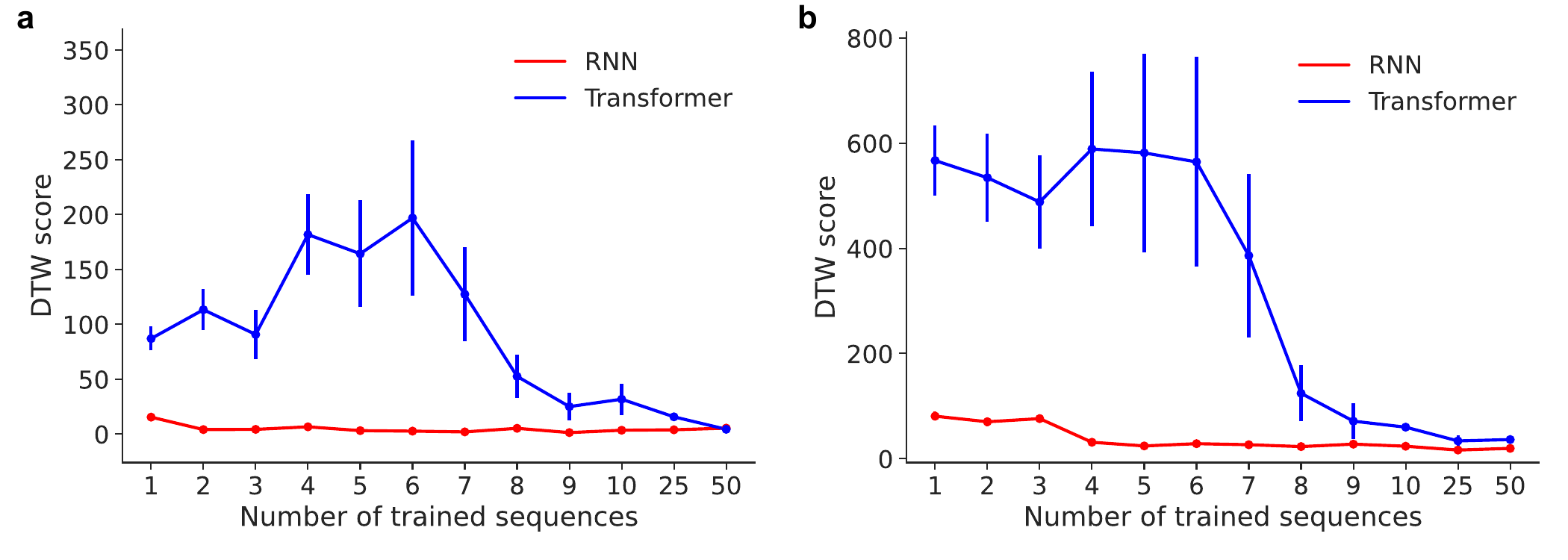}
    \caption{\textbf{Quantitative evaluation of model performance using Dynamic Time Warping (DTW).} \textbf{a, b,} The figure presents the average DTW plot along with the standard error for models trained with point and circle attractors, respectively.}
    \label{fig:dtw_score}
\end{figure}

\begin{figure}[t]
    \includegraphics[scale = 0.28]{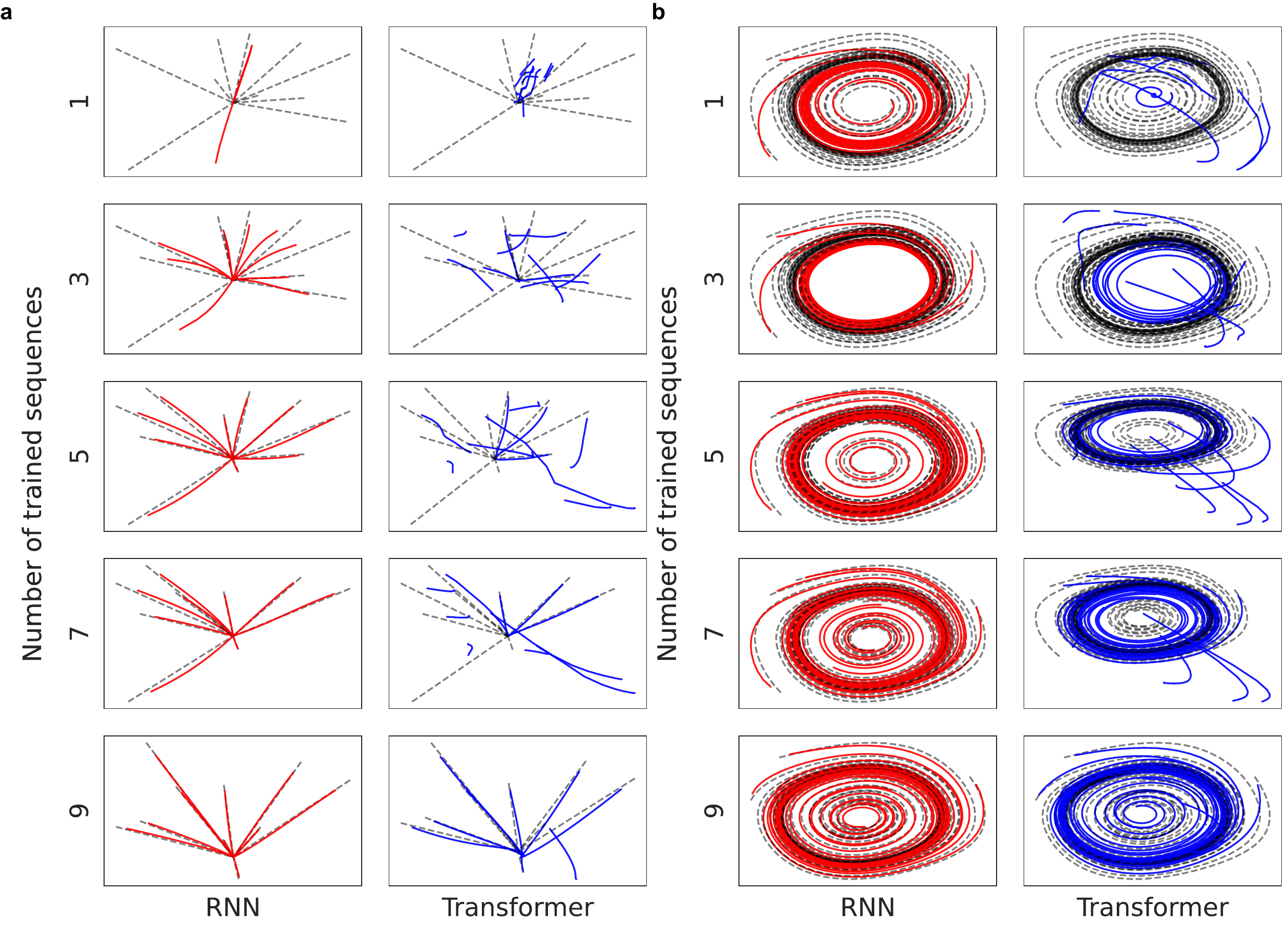}
    \caption{\textbf{Qualitative evaluation of model performance.} \textbf{a, b,} The outputs of trained models using point attractor and cyclic attractor dynamics, respectively, employ varying numbers of sequences. Colored lines represent the outputs, while the true trajectory corresponding to the provided initial position is depicted using dot lines.}
    \label{fig:output}
\end{figure}

The ability to extract essential structures underlying a limited number of experiences is a core AI capability. This experiment was designed to explore the capacity of generalization in learning (GIL) in both RNN and Transformer models. Models were trained using two forms of attractor dynamics, i.e., point and cyclic attractors. We systematically varied the number of training sequences to examine the quantity of training data necessary to extract essential structure underlying a dataset (see Methods; \textbf{trajectories of attractor dynamics}). Given a random initial position denoted by xy coordinates, the models were required to predict the position one step ahead in an autoregressive manner for T time steps. For performance evaluation, the Dynamic Time Warping (DTW) method has been employed (see Methods; \textbf{evaluation settings}).

These experimental results showed that the RNN successfully acquired attractor dynamics using significantly fewer data than the Transformer model. Interestingly, as illustrated in Figure \ref{fig:dtw_score}, the RNN converged to almost the correct trajectory with only 2-4 training datasets, whereas the Transformer model required approximately 25-50 datasets to achieve similar performance (for more detail, see Appendix; Table \ref{table:DTW_score_point_attractor} and \ref{table:DTW_score_cyclic_attractor}). In Figure \ref{fig:output}, a qualitative evaluation reveals that the RNN already achieved a highly satisfactory representation with just three sequences. Notably, in the realm of cyclic attractors, the RNN model produced output that converged upon a limit cycle, even when trained on a single sequence. Conversely, the Transformer model struggled to reproduce attractors under the same initial conditions, despite being trained with nine sequences. It is important to note that we conducted the same experiments involving varying dropout rates to address concerns related to the Transformer's susceptibility to overfitting, and the results remained consistent (see Appendix; \textbf{Model variations to avoid overfitting}). Furthermore, we extended our comparative analysis to address the inherently more complex task of human hand-drawn pattern learning (see Appendix; \textbf{Human hand-drawn pattern learning}).

\section*{Discussion}
This study investigated the difference in generalization in learning (GIL) between RNN and Transformer. Our results revealed that the RNN model has the capacity to extrapolate fundamental dynamic structure of attractors from even a limited number of data, in contrast to the Transformer model, which requires considerably more data for the same task. These results support our hypothesis that generalization by Transformer is considerably weaker when working with limited numbers of exemplars.

Compared to recurrent models, the deficiency of feed-forward models, which are utilized in the Transformer model, has been reported in various tasks \cite{Karim1992, Sundermeyer2013, Alamia2020}. This shortfall in feed-forward models is often attributed to their inability to acquire ``recurrent" inductive bias \cite{Tran2018, Dehghani2018}. From a different perspective, Elman et al. \cite{Elman1990, Servan1988, Tani2003, Namikawa2008} have demonstrated that distributed representation in neural networks can yield a global structure that accounts for both learned and unlearned patterns, while local representation \cite{Wolpert1998, Tani1999, Haruno2001} % which allocates specific tasks or information processing among modules 
is inferior in its capacity to capture global data structure \cite{Elman1996, Mcclelland1987}. Given reports, we attribute the dissimilarity of the GILs observed in this study to the ability to acquire ``recurrent" inductive bias inherent to shared weights.

Given that the artificial intelligence community has overwhelmingly favored scaling as the primary method to improve performance, the study of GIL with limited exemplars has been somewhat overlooked \cite{Hoffmann2022, Frank2023}. Nevertheless, this study reveals substantial disparities in GIL when working with limited data. Our findings have important implications and underscore potential relevance for architecture selection in practical applications where data scarcity is a common constraint, such as medicine, finance, and robotics. We acknowledge there are several limitations to this study, notably that our experiment uses only the vanilla Transformer-based model. Hence, further investigations employing recently proposed models will be conducted \cite{Dehghani2018, Wang2019, Bulatov2022}, which employs the ``recurrent" inductive bias. We hope that this work will motivate others to further investigate GIL with limited exemplars and that these insights will be useful in developing the next generation of neural networks.

\section*{METHODS}
\begin{figure}[t]
    \includegraphics[scale = 0.4]{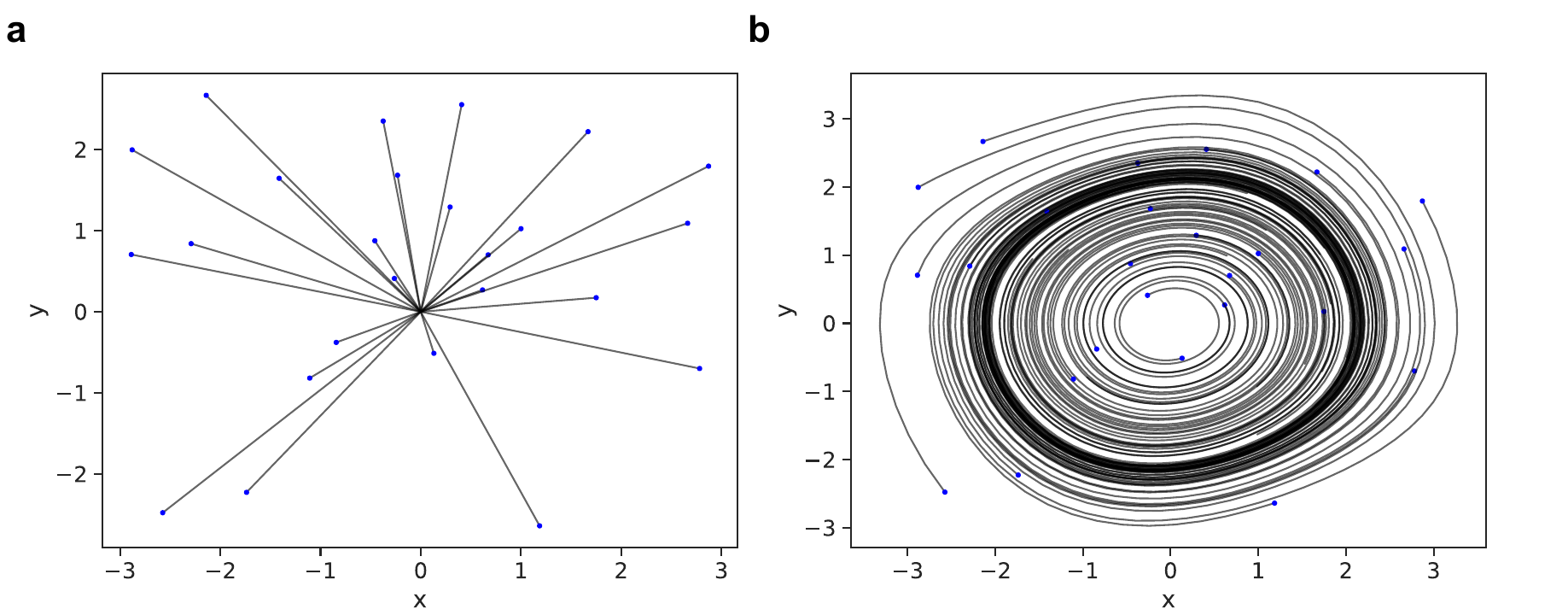}
    \caption{\textbf{Trajectories of Attractor Dynamics.} 
    \textbf{a, b,} Examples of point and cyclic attractor dynamics, respectively, with 25 random initial positions. These trajectories were generated based on Equations 1 and 2, respectively. The initial positions are denoted by blue dots.}
    \label{fig:train_data}
\end{figure}

\textbf{Trajectories of attractor dynamics.} Since our primary focus was to evaluate the capacity of GIL in the models, two relatively simple types of attractor dynamics, i.e., point and cyclic attractors, were used as training trajectories. Each trajectory was initiated from random initial positions in the xy-coordinate range of [-3.0, 3.0] and transited for T time steps. Point attractors were generated utilizing equations \ref{eq:point_attractor} with the parameter $\alpha$ set to -1.0 and T set to 100. This equation lets the point at a random initial position move linearly closer to the fixed point attractor located at [0, 0] with gradual deceleration. Cyclic attractors, on the other hand, were generated based on the van der Pol oscillator, shown in equation \ref{eq:cyclic_attractor}, when the parameter $\mu$ was set to 0.1, and T was set to 200. Figure \ref{fig:train_data} illustrates examples of trajectories corresponding to point and cyclic attractors, each initiated from 25 random initial values.

\begin{equation}
\label{eq:point_attractor}
\begin{cases}
\frac{dx}{dt} = ax,\\
\frac{dy}{dt} = -y %\nonumber
\end{cases}
\end{equation}

\begin{equation}
\label{eq:cyclic_attractor}
\begin{cases}
\frac{d^2x}{dt^2} - \mu(1-x^2)\frac{dx}{dt} + x = 0,\\
y = \frac{dx}{dt}\end{cases}
\end{equation}

\textbf{RNN model.} The RNN model comprises two layers: an Elman recurrent layer and an output layer. The Elman recurrent layer consists of 20 hidden units with a tanh activation function. In the training phase, the initial hidden state is initialized as a trainable parameter, allowing it to adapt to each of the different trajectories. In the evaluation phase, it is initialized to a value of 0 \cite{Nishimoto2008}.

\textbf{Transformer model.} The Transformer model used in this study is a decoder-only sequence model. This is a well-used architecture to generate the next words/actions in the sequence in language/action modeling, such as GPT-based models or Robotic Transformer \cite{Radford2018, Brohan2022}.
The Transformer model comprises four layers: an embedding layer, a positional encoding layer, a Transformer decoder layer, and an output layer. The embedding layer consists of 20 hidden units, and the Transformer decoder layer employs 40 units with four multi-heads attention. For the positional encoding layer (PE), we use the trigonometric functions proposed in \cite{Vaswani2017} formulated as:

\begin{equation}
\label{eq:position_encoding}
\begin{cases}
\mathit{PE}_{(\mathit{pos},2i)} = \sin\left(\frac{pos}{10000^{2i/d_{\mathit{model}}}}\right)\\
\mathit{PE}_{(\mathit{pos},2i+1)} = \cos\left( \frac{pos}{10000^{2i/d_{\mathit{model}}}} \right),
\end{cases}
\end{equation}
where, $pos$ is the position, $i$ is the dimension, and $d_{\mathit{model}}$ has the same dimensions as the embedding layer. During the training phase, a masked self-attention technique was employed in the decoder to prevent attending to subsequent positions in the sequence.

\textbf{Training settings.} The networks are trained in a supervised manner by minimizing the mean squared error between the network prediction and the next position. The Adam optimizer \cite{Kingma2014} with $\alpha=0.001, \beta_{1}=0.9, \beta_{2}=0.999, and \epsilon=10^{-8}$ has been used for the implementation. The networks are trained for 25,000 epochs using NVIDIA GeForce RTX 2080 Ti GPU. 

\textbf{Evaluation settings.} We assessed network performance on two types of attractor dynamics by giving 10 random initial positions. Networks were required to autoregressively generate subsequent trajectories based on the given initial values. To quantitatively evaluate the quality of trajectory generation, we adopted the Dynamic Time Warping (DTW) technique, which is widely recognized for its effectiveness in addressing temporal distortions and shifts in time series data across various applications. It is important to note that a lower DTW score indicates greater similarity between the two time-series sequences, meaning that they align more closely or exhibit fewer distortions. We computed the DTW distance between trajectories generated by these networks and the true trajectories corresponding to the provided initial positions. Thus, a model with a smaller DTW implies a superior resemblance between network-generated trajectories and the true trajectories. We computed DTW distances between trajectories generated by the networks and the true trajectories corresponding to the provided initial positions. Thus, a model exhibiting a smaller DTW score implies superior performance in extracting fundamental structure inherent to a dataset.
\bibliographystyle{naturemag} % bibstyle for the Nature journal
\bibliography{references}

\section*{Appendix}
\textbf{Model variations to avoid overfitting.} 
The same experiments were performed by incorporating dropout layers and random masks into the Transformer model. Dropout \cite{Srivastava2014} is a regularization technique commonly used in neural networks to prevent overfitting and to enhance the model's capacity for generalization. Specifically, Transformer-based models with a substantial number of parameters have demonstrated better performance by utilizing dropout layers. Thus, despite the relatively modest number of parameters in our Transformer model, we pursued an additional experiment by introducing dropout layers. As in \cite{Vaswani2017}, we apply dropout to the output of each sub-layer and the sums of the embeddings and the positional encodings. The base model uses a dropout rate of $P_{drop}=0.0$.

\begin{figure}[th]
    \includegraphics[scale = 0.4]{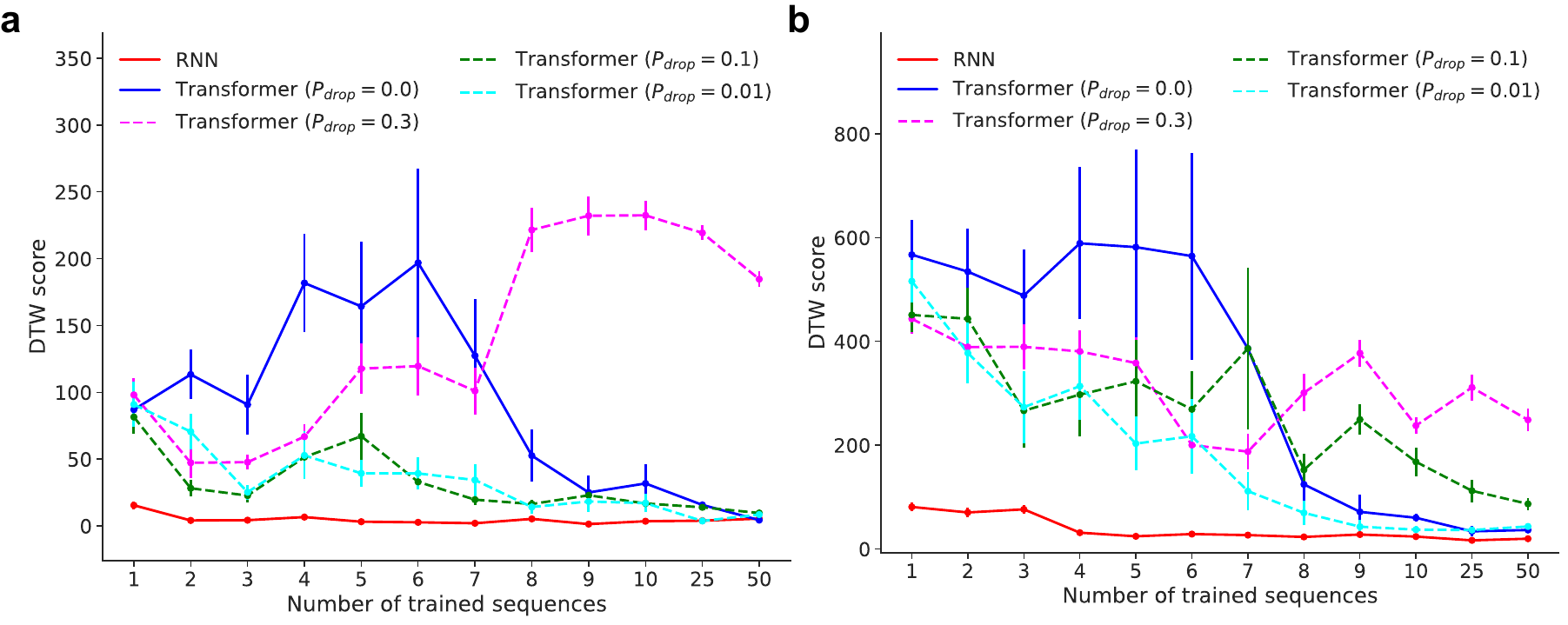}
    \caption{\textbf{Model variation study.} \textbf{a,b,} The figure presents the average DTW plot along with the standard error for models trained with point and circle attractors, respectively.}
    \label{fig:dtw_dropout}
\end{figure}

Figure \ref{fig:dtw_dropout} illustrates the average Dynamic Time Warping (DTW) scores for model outputs across various amounts of training sequences. Transformer models equipped with an appropriately fine-tuned dropout rate ($P_{drop}=0.01$) demonstrate superior performance compared to models without dropout layers, particularly when dealing with limited training datasets. However, in contrast to recurrent neural network (RNN) models, transformer models have yet to attain a satisfactory level of representation, achieving scores closely aligned with RNNs when training with approximately 25-50 sequences for point attractors and 10 sequences for cyclic attractors. Also, models with $P_{drop}=0.1$ and $P_{drop}=0.3$ exhibit underfitting behavior within this setting. For more detailed scores, see Table \ref{table:DTW_score_point_attractor} and \ref{table:DTW_score_cyclic_attractor}. 

We also conducted experiments with a combination of random masks. Training with random masks enables different training objectives and combinations, so that more training signals can be extracted from a given trajectory, and thus the model becomes more data efficient \cite{Li2023, Wu2023}. However, the conclusions remained the same, and hence, the results are omitted here.

% Training with random masks enables different training objectives or combinations, thus allowing more learning signals to be extracted from any given trajectory. As a result, model could be more data efficient \cite{Li2023, Wu2023}

\begin{table}[b]
  \caption{DTW score with standard error. Models trained with point attractor}
  \label{table:DTW_score_point_attractor}
  \hspace{-1.0cm}
  \begin{tabular}{c||cc||ccc}
    \hline 
    \multirow{2}{*}{\shortstack[c]{Number of \\trained sequences}}    & \multicolumn{5}{c}{Model} \\
    & RNN & Transformer &$P_{drop}=0.3$ & $P_{drop}=0.1$ & $P_{drop}=0.01$\\
    \hline \hline
    \rowcolor{lightgray}1 & 15.4$\pm$3.1 & 87.0$\pm$10.9 & 98.0$\pm$12.7 & 81.5$\pm$12.3 & 90.8$\pm$16.9\\
    2 & 4.0$\pm$0.8 & 113.3$\pm$18.6 & 47.2$\pm$11.5 & 28.1$\pm$6.1 & 70.5$\pm$13.4\\
    \rowcolor{lightgray}3 & 4.2$\pm$0.9 & 90.7$\pm$22.5 & 47.7$\pm$5.7 & 22.7$\pm$5.1 & 25.2$\pm$5.4\\
    4 & 6.5$\pm$1.0 & 181.7$\pm$36.8 & 66.8$\pm$9.2 & 51.4$\pm$8.8 & 52.7$\pm$17.9\\
    \rowcolor{lightgray}5 & 3.1$\pm$0.3 & 164.3$\pm$48.7 & 117.7$\pm$21.9 & 67.1$\pm$17.4 & 39.3$\pm$10.1\\
    6 & 2.6$\pm$0.3 & 196.9$\pm$70.6 & 119.6$\pm$21.9 & 32.9$\pm$5.9 & 39.2$\pm$12.0\\
    \rowcolor{lightgray}7 & 1.9$\pm$0.4 & 127.3$\pm$42.7 & 101.0$\pm$17.5 & 19.6$\pm$3.9 & 34.3$\pm$12.0\\
    8 & 5.2$\pm$0.8 & 52.5$\pm$19.7 & 221.5$\pm$16.8 & 16.3$\pm$3.0 & 14.1$\pm$4.7\\
    \rowcolor{lightgray}9 & 1.3$\pm$0.1 & 25.0$\pm$12.5 &232.1$\pm$14.5 & 22.9$\pm$4.2 & 18.2$\pm$7.6\\
    10 & 3.4$\pm$0.1 & 31.7$\pm$14.3 & 232.4$\pm$11.2 & 16.7$\pm$1.2 & 17.0$\pm$6.9\\
    \rowcolor{lightgray}25 & 3.8$\pm$0.1 & 15.7$\pm$0.8 & 219.3$\pm$5.6 & 13.9$\pm$0.5 & 3.6$\pm$0.3\\
    50 & 5.4$\pm$0.3 & 4.3$\pm$0.3 & 184.7$\pm$5.8 & 9.6$\pm$1.3 & 8.4$\pm$0.6\\
    \hline
  \end{tabular}
\end{table}

\begin{table}[t]
  \caption{DTW score with standard error. Models trained with cyclic attractor}
  \label{table:DTW_score_cyclic_attractor}
  \hspace{-1.0cm}
  \begin{tabular}{c||cc||ccc}
    \hline 
    \multirow{2}{*}{\shortstack[c]{Number of \\trained sequences}}    & \multicolumn{5}{c}{Model} \\
    & RNN & Transformer &$P_{drop}=0.3$ & $P_{drop}=0.1$ & $P_{drop}=0.01$\\
    \hline \hline
    \rowcolor{lightgray}1 & 80.8$\pm$8.7 & 567.0$\pm$67.0 &443.4$\pm$29.3 & 450.9$\pm$33.5 & 515.9$\pm$41.3\\
    2 & 70.2$\pm$8.5 & 534.3$\pm$83.8 & 389.1$\pm$42.7 & 443.4$\pm$59.9 & 377.3$\pm$58.2\\
    \rowcolor{lightgray}3 & 76.2$\pm$7.8 & 488.2$\pm$88.6 & 389.4$\pm$43.0 & 266.4$\pm$71.4 & 273.1$\pm$69.3\\
    4 & 31.2$\pm$4.2 & 589.0$\pm$146.6 & 380.6$\pm$40.0 & 297.3$\pm$80.3 & 313.6$\pm$64.4\\
    \rowcolor{lightgray}5 & 24.2$\pm$2.4 & 581.5$\pm$188.7 & 358.2$\pm$50.5 & 322.9$\pm$79.6 & 203.0$\pm$51.8\\
    6 & 28.5$\pm$3.4 & 564.3$\pm$199.3 & 200.1$\pm$39.2 & 269.0$\pm$73.4 & 217.0$\pm$71.5\\
    \rowcolor{lightgray}7 & 26.5$\pm$3.3 & 386.1$\pm$155.8 & 187.4$\pm$34.9 & 386.1$\pm$155.8 & 111.3$\pm$37.2\\
    8 & 23.0$\pm$1.5 & 124.3$\pm$53.6 & 300.8$\pm$36.0 & 152.4$\pm$31.4 & 69.4$\pm$23.7\\
    \rowcolor{lightgray}9 & 27.6$\pm$3.3 & 71.4$\pm$33.5 & 377.1$\pm$26.1 & 249.3$\pm$29.7 & 42.8$\pm$13.0\\
    10 & 23.7$\pm$2.7 & 60.0$\pm$7.6 & 237.6$\pm$15.5 & 167.8$\pm$27.5 & 37.0$\pm$6.7\\
    \rowcolor{lightgray}25 & 16.4$\pm$1.2 & 33.8$\pm$10.2 & 311.1$\pm$25.0 & 111.9$\pm$21.9 & 36.3$\pm$5.5\\
    50 & 19.6$\pm$1.4 & 36.4$\pm$2.4 & 248.6$\pm$21.1 & 86.7$\pm$12.0 & 43.2$\pm$4.5\\
    \hline
  \end{tabular}
\end{table}

\begin{figure}[t]
    \includegraphics[scale = 0.32]{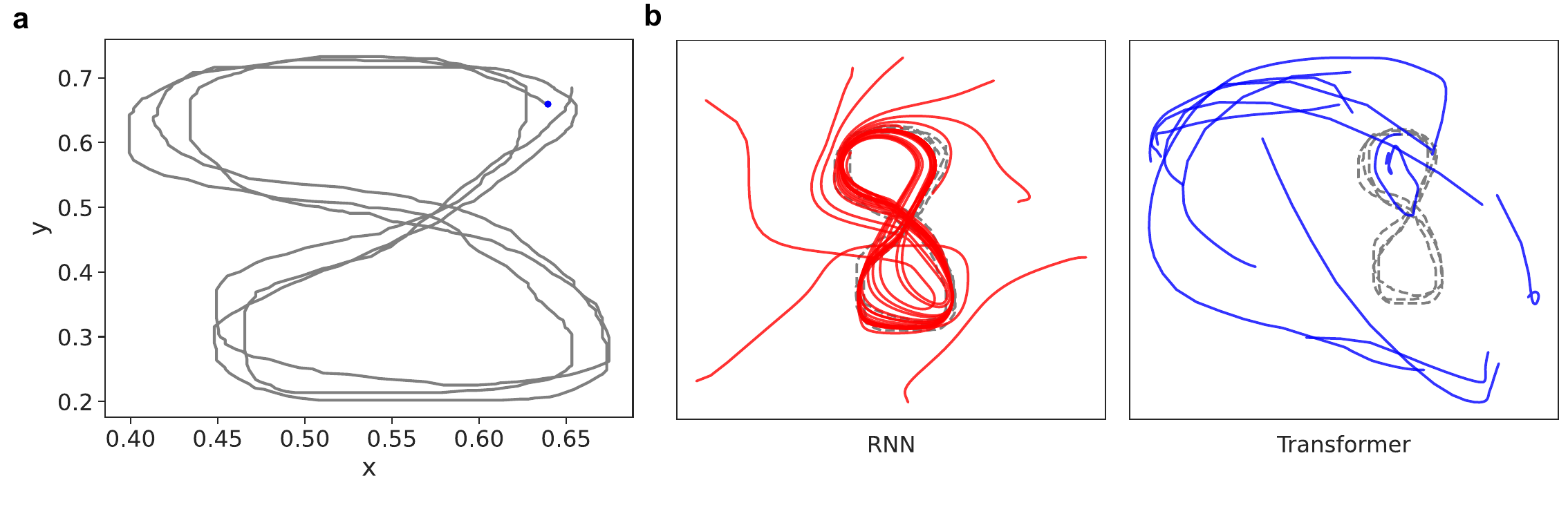}
    \caption{\textbf{Result of a human hand-drawn pattern learning.} \textbf{a,} Training data. The trajectory is drawn by a human. Blue dot points denote the initial positions. \textbf{b,} Model output from ten random initial positions. Black dot lines indicate the trajectory of training data.}
    \label{fig:figure8}
\end{figure}

\textbf{Human hand-drawn pattern learning.} 
Given the inherent variability in amplitude, velocity, and shape of human hand-drawn patterns, the generated data is inherently more complex compared to the trajectories employed in the aforementioned study. For this particular investigation, a singular training trajectory was employed (see Figure \ref{fig:figure8}; \textbf{a}). This trajectory traced a figure-eight pattern approximately three times, encompassing a total of 560 time steps.

Figure \ref{fig:figure8}; \textbf{b} shows that the RNN was able to extract the underlying essential structure, i.e., converging to the figure 8 attractor, while the Transformer model struggles to extract meaningful patterns.

% \textbf{Code availability statement.} Our networks are implemented by using PyTorch. We will release the code for this experiment as a GitHub repository.
\textbf{Data availability statement.} The datasets used during the current study are available from the corresponding author (R.F.) upon request. The codes for this experiment, including the data generation, will also be made available as a GitHub repository.

\textbf{Author contributions} J.T. conceived the concept. R.F. conducted the experiments and wrote the manuscript.

\textbf{Competing interests declaration} The authors declare that the research was conducted in the absence of any commercial or financial relationships that could be construed as a potential conflict of interest.

\end{document}